\RequirePackage{fix-cm}
\documentclass[smallextended]{svjour3}       % onecolumn (second format)
\smartqed  % flush right qed marks, e.g. at end of proof
\usepackage{amsmath}
\usepackage{graphicx}
\usepackage{setspace}
\usepackage[T1]{fontenc}
\usepackage{bm}
\usepackage{lineno}
\usepackage{multirow}
\usepackage{xcolor}
\usepackage{amsmath}
\usepackage{caption}
\usepackage{tabularx}

\usepackage[linesnumbered,ruled,vlined]{algorithm2e}
\usepackage{comment}
\DeclareCaptionFont{blue}{}
\DeclareCaptionFont{black}{}
\newcommand{\name}{{\sc PatternCausality}}
\newcommand{\test}{{\sc MedCause}}

\begin{document}
\title{Informative Causality Extraction from Medical Literature via Dependency-tree based Patterns}

%\author[1]{M Ahsanul Kabir}
%\author[2]{AlJohara Almulhim}
%\author[3]{Xiao Luo}
%\author[4]{Mohammad Al Hasan}

%\affil[1,2,4]{Department of Computer Science, Indiana University Purdue University Indianapolis, USA, {\bf Email:} mdkabir@iu.edu, aalmulhi@iu.edu, alhasan@iupui.edu}
%\affil[2]{Department of Computer and Information Technology, Indiana University Purdue University Indianapolis, USA, i{\bf Email:} luo25@iupui.edu}

\author{M. Ahsanul Kabir \and AlJohara Almulhim \and Xiao Luo \and Mohammad Al Hasan\footnote{Corresponding Author}}

%\authorrunning{Short form of author list} % if too long for running head

\institute{M. Hasan \at Indiana University Purdue University Indianapolis \newline
           Dept. of Computer Science, Indianapolis, IN, USA \newline
           \email{alhasan@iupui.edu}, Phone: +1-317-460-2438
}

\maketitle
%\onehalfspacing
%\linenumbers
\begin{abstract}

Extracting cause-effect entities from medical literature is an important task in medical information  retrieval. A solution for solving this task can be used for compilation of various causality relations, such as, causality between disease and symptoms, between medications and side-effects, between  genes and diseases, etc. Existing solutions for extracting cause-effect entities work well for sentences where the cause and the effect phrases are name entities, single-word nouns, or noun phrases consisting of  two to three words. Unfortunately, in medical literature, cause and effect phrases in a sentence are  not simply nouns or noun phrases, rather they are complex phrases consisting of several words, and  existing methods fail to correctly extract the cause and effect entities in such sentences. Partial  extraction of cause and effect entities  conveys poor quality, non-informative, and  often, contradictory facts, comparing to the one intended in the given sentence. In this work, we solve this problem by designing an unsupervised method for cause and effect phrase extraction, \name, which is specifically suitable for the medical literature. Our proposed approach first uses  a collection of cause-effect dependency patterns as template to extract head words of cause and  effect phrases and then it uses a novel phrase extraction method to obtain complete and meaningful cause and effect phrases from a sentence. Experiments on a cause-effect dataset built from sentences  from PubMed articles show that for extracting cause and effect entities, \name\ is substantially  better than the existing methods---with an order of magnitude improvement in the F-score metric over the best  of the existing methods. We also build different variants of \name, which use different phrase extraction methods; all variants are better than the existing methods. \name\ and its variants also show modest  performance improvement over the existing methods for extracting cause and effect entities in a  domain-neutral benchmark dataset, in which cause and effect entities are nouns or noun phrases consisting of one to two words.

\end{abstract}

\section*{Keywords}
Causality, Cause-effect phrase extraction, Cause-effect pattern

\section{Introduction}

Medical research is advancing at a rapid pace causing information overload in the
medical literature. To overcome this challenge and to be up-to-date about the latest discoveries, 
medical communities extract health-related facts, hypotheses, and real-life evidences 
from various textual sources, including scientific reports, clinical notes, and online 
health forums. For this purpose they use various text mining and information extraction 
methodologies. Among these, causality extraction is one of the most important extraction 
methodologies due to the fact that causal sentences are generally used for describing
medical facts and hypotheses.
For instances, causal sentences are commonly used in medical text for providing the causes of a 
disease. For instance, the sentence , ``diabetes is caused 
by the absence of insulin secretion, and vitamin D deficiency'', reflects disease causality, clearly denoting two of the causes of
the diabetes disease. 
Effectively extracting disease causality 
\cite{David:2002:causality,Tal:2000:Disease-Causality,Walter:2020:Disease-Causality}
from such sentences provides knowledge which may help prevention of that disease.
Causal sentences can also help discovering the effectiveness of a medical 
treatment or the side effects associated with the treatment~\cite{Evans:2003:drug-effects,Berry:2002:drug-effects}. For example, the sentence, ``ACE Inhibitors are 
used to treat heart-related conditions, but can also cause chronic 
cough''~\cite{Peter:2006:cough}, reflects side effect causality. From such sentences if the
cause and the effect phrases can be extracted automatically, researchers can derive
facts, build relevant hypotheses, and enrich the medical knowledge-base effectively and efficiently. 

Due to the importance of causality extraction in medical domains, several
works~\cite{atkinson2008discovering,lee2017disease,zhao2018causaltriad,an2019extracting} have
been proposed, which either identify causal relations from sentences or extract cause and 
effect pairs to build causal networks. These existing works either use a set of known 
syntactic patterns to extract the cause and effect entities or use machine learning models to 
identify those entity pairs. Considering the cause and effect pairs as name entities, existing 
methods focus on entity extraction, and they performed well when the causes and effects are 
name entities, or noun phrases, such as, the name of diseases, medications, or genes. 
However, in medical text, the cause or the effect phrases can often be complex, including not 
only a name entity but also a multi-word phrase consisting of noun or pronouns preceding with 
modifiers, or even a dependent clause. In those scenarios, the existing methods fail to extract 
complete phrases,  often returning only head words, which is not enough to capture the intended 
causality. In fact, such a shortcoming can, sometimes, alter the intended 
causality information completely. For example, consider the sentence ``Deficiency in 
Vitamin D can cause increased mortality rate in Covid-19 patients''. One of the existing methods,
which we refer as Logical-Rule Based
method~\cite{sorgente2013automatic} extracts `Vitamin D' 
as the  cause term and `increased mortality rate'  as the effect term, seemingly conveying 
the erroneous fact 
that ``Vitamin D causes increase mortality rate'---whereas, it is not ``Vitamin D'', but the 
``deficiency of Vitamin D'' is the culprit. Another existing method which we refer as Word Vector Mapping Based method~\cite{xiao:2019:graph} extract ``Defficiency'' as the cause term and ``increase mortality 
rate'' as the effect term, leaves the readers dumbfounded with the question, deficiency of what?
Given the abundance of longer cause and effect phrases in the medical text, a new 
causality extraction method is needed which is better at capturing the complete cause and effect 
phrases, thus retaining the integrity of the fact intended in a causal sentence.

To overcome this difficulty, in this work we propose a simple, yet highly effective, causality 
extraction method, which is particularly suited for extracting causality from scientific documents, 
where the cause and effect terms are generally longer. Our proposed method uses a dependency tree 
parser and then utilizes a collection of cause-effect dependency patterns for extracting the cause 
and effect nodes in the dependency tree of a sentence. Then it uses a novel phrase extraction 
method to obtain complete cause and effect phrases from that sentence.
Experiments on a corpus built from PubMed articles shows that \name\ 
can successfully extract long cause and effect phrases whereas existing methods fail to do 
so. As for the previous example, \name\ is able to extract `deficiency in Vitamin D' as the 
cause term of the given sentence.

\section{Related Work}

Causal relation extraction tasks can be mainly categorized into unsupervised~\cite{information.retrieval.1998,information.retrieval.2001}, supervised~\cite{dasgupta-etal-2018-automatic,bi-lstm-crf-2019}, and hybrid approaches~\cite{question-answering-2006,Antonio:2013:baseline,xiao:2019:graph}. The unsupervised approaches are mainly pattern based approach, which use causative verbs, causal links, causal relations between words or phrases to extract cause effect pairs. The supervised approaches need to have a training set with cause and effect phrase pairs labeled, then supervised learning models can be trained to extract the causal relationships between the phrases. Do et al. \cite{do2011minimally} developed a minimal supervised approach based on constrained conditional model framework with an objective function that takes 
discourse connectives into consideration. Dasgupta et al. \cite{dasgupta-etal-2018-automatic} used word embedding with selected linguistic features to construct the representation of entities as input of a bidirectional Long-Short Term Memory (LSTM) model to predict causal entity pairs. Nguyen et al. \cite{nguyen2015relation} utilized pre-trained word embedding to train convolution Nerual Network (CNN) to classify given casual pairs. Peng et al. \cite{peng2017cross} presented a model-based approach that utilizes deep learning architecture to classify relations between pairs of drugs and mutations, and also triplets of drugs, genes and mutations with N-ary relations across multiple sentences extracted from PubMed corpus. %They initialize given entities with GLoVe representations\cite{pennington2014glove} ,then use Long Short-Term Memory(GraphLSTM) to get meaningful contextual representations, later they use these vectors as features to train number of logistic regression classifiers for each extracted relation. 
Zhao et al. \cite{zhao2018causaltriad} developed a semi-supervised approach which leverages both contextual information and graph structure to extract unseen causal pairs among multiple sentences within medical text. The known causal pairs were determined using context vector that is calculated conducting TFIDF on concatenating the words before, after and between the causal pairs, and the differences between the causal pair embeddings that is generated using Word2Vec. %Selecting triads among a window of sentences assure the process of learning and relations discovery of pairs which are likely share the same context unlike making arbitrary selections. Cause-effect relations of medical detests are studied over six defined types/pattern. First, contextual information of trained triads is modeled using two methods: synthetic symbolic context and word embedding. Then, a factor graph is constructed using partially annotated triads to model transitivity between pairs where casual relations are discovered based on the predefined factors.
The hybrid approaches use both patterns as well as supervised models to extract causality. One of the earliest hybrid approaches is proposed by Sorgente \cite{sorgente2013automatic} which has two phases: lexico-syntactic patterns extraction and machine learning model classification.  %first phase, it extracts causal patterns based on words dependencies in a given sentence over four set of rules with defined regular expressions. In the second phase, the extracted pairs are fed into binary classification Bayesian model to extract true cause-effect pairs and Laplace smoothing to filter noncausal pairs. %Each pair of entities are trained using a combination of lexical and dependency features extracted by selecting all homonyms and synonyms similar to the given two entities which are found in wordNet \cite{christiane1998wordnet} and from direct connection of entities in a parse tree, respectively. Another approach has been provided as a python-based keyphrase extraction toolkit known as pke. \cite{boudin:2016:COLINGDEMO} This tool has two options to train candidate pairs extracted from documents: unsupervised and supervised implementations. In unsupervised mode, candidate pairs are fed into weighted function to select top-N pairs based on their highest weights. Functions used to rank candidates includes: TfIdf\cite{kim2010semeval}, SingleRank\cite{wan2008collabrank}, TopicRank\cite{bougouin2013topicrank}, and KP-Miner\cite{el2010kp}. On the other hands, the supervised mode defines a a naive bayes classifier to train pairs with a range of document logical structure information features such as: term frequency known as TF and IDF defined as inverse document frequency in order to filter out pairs based on their confidence scores.

Text mining and machine learning approaches have been employed in the medical domain for causal 
relation extraction. Atkinson et al. \cite{atkinson2008discovering} worked on biomedical texts 
causal pattern discovery using Bayesian net, but the main focuse of this work is causal relation classification; for instance, classifying between harmful and beneficial causal relations. Lee et al. \cite{lee2017disease} investigated disease causality extraction using lexicon-based causality term strength and frequency-based causality strength. In this work, the causality relations are defined 
by a set of terms, such as, ``causes'', ``affects'', etc with associated strength. An et al. \cite{an2019extracting} explored extracting causal relations using syntactic analysis with word embedding. They used designed syntactic patterns to first obtain triples including the cause and effect pairs, verb that links the pairs and a binary term denoting whether the relation is passive or negative.  Then,
word embedding of the verbs in the designed syntactic patterns were used to discovery additional verbs that define the causal relations.  

%\subsection*{Objective}
%Both supervised and hybrid approaches require a set of training data to train a model for casual pairs extraction, which focus on classifying a pair of phrases to casual relations. The syntactic patterns of the casual relationships cannot be extracted by the supervised or hybrid approaches. They often failed to capture complex causal pairs if their frequency is low in the training data, whereas complex and long casual pairs present commonly in the medical context. For example, in sentence ``'', the casual pairs are ``'' as cause and ``'' as effect. In addition, one sentence in a medical literature or text may contain multiple  causal pairs. For example, ..... Therefore, the objective this work is to develop a unsupervised learning framework to detect and extract the multiple and complex casual pairs from medical literature. Meanwhile, the syntactic patterns of the casual relationship identified by the unsupervised learning can be applied to other medical text to extract casual pairs.

%Our main objective of this work is to extract the complete cause and effect term and thus to maximize cause effect information extraction from a sentence. We want to provide a software which can extract the cause effect terms from a sentence automatically.

\section{Materials and Methods}
Given a sentence $\mathcal{S}$, which denotes causality between two phrases, 
hereby called as cause phrase, $u$ and effect phrase, $w$, our task is to extract the phrases $u$ and $w$, as completely as possible. We assume that the sentences are from scientific literature where the cause and effect phrases are generally longer. To extract the cause and effect phrases we use a collection of cause-effect dependency patterns (will be defined later), denoted as $\mathcal{P}$. These 
patterns are used as template to extract $u$ and $w$. 

Given a sentence, we first find all its noun phrases by dependency tree parsing. Each
of these phrases is a candidate to be the cause phrase or the effect phrase. For
every such pair of noun phrases, we then validate whether the pair of phrases fits 
with one of the dependency patterns. If yes, we extract the phrases as potential
cause and effect phrases. However, these phrases may not be complete, rather they
could merely be head words of a complete cause or effect phrase; so we extend
these phrases by finding sentence segments which are associated with the descendant
nodes of the cause and effect phrase nodes in the dependency tree. The main novelty of \name\ is two-fold: First, the
utilization of a collection of dependency patterns for extracting core part of cause 
and effect phrases; Second, the extension of the core part of cause and effect
phrases by finding sentence segments associated with the correct dependency tree 
nodes. 

Below we discuss the entire process in four different steps: sentence representation, dependency pattern extraction, pattern matching and core cause-effect phrase extraction, and cause-effect 
phrase extension.

\subsection{Sentence Representation}
To identify cause effect phrases from a sentence, the sentence should be
represented in a form that can preserve its syntactic structure. In such
a representation, the syntactic units in a sentence are isolated, which
would make the extraction of the cause and effect phrases easier.
Dependency tree parsers, such as, Spacy ~\cite{Spacy2}, Stanford~\cite{stanford_nlp_2008}, ClearNLP~\cite{choi-palmer-2012-fast}, LTH~\cite{johansson-nugues-2007-extended}, etc.\ can serve this purpose.
The output of such a parser is a tree in which each node
corresponds to a word or a phrase denoting a syntactical unit. The nodes
are also labeled by the parts-of-speech of the word or phrase associated to a 
node. Dependency among different syntactic units are reflected by tree edges,
which are labeled by dependency relations. While most of the dependency parsers generate more or less similar dependency tree, we choose Spacy~\cite{Spacy2} for
this task due to the following reasons: First, Spacy provides an industrial strength API 
for broader natural language processing tasks, allowing us to build an end-to-end NLP
application, including libraries for tokenization, data import, and visualization, in
addition to libraries for building  the dependency parser; Second, Spacy is well 
documented and can easily be used for the dependency tree extraction task. Finally, 
Spacy is very efficient in terms of execution time.
For all of the experiments and examples of this research, we have used Spacy for dependency parsing. 

\begin{figure*}[h!]
    \centering
    \includegraphics[width=1\linewidth]{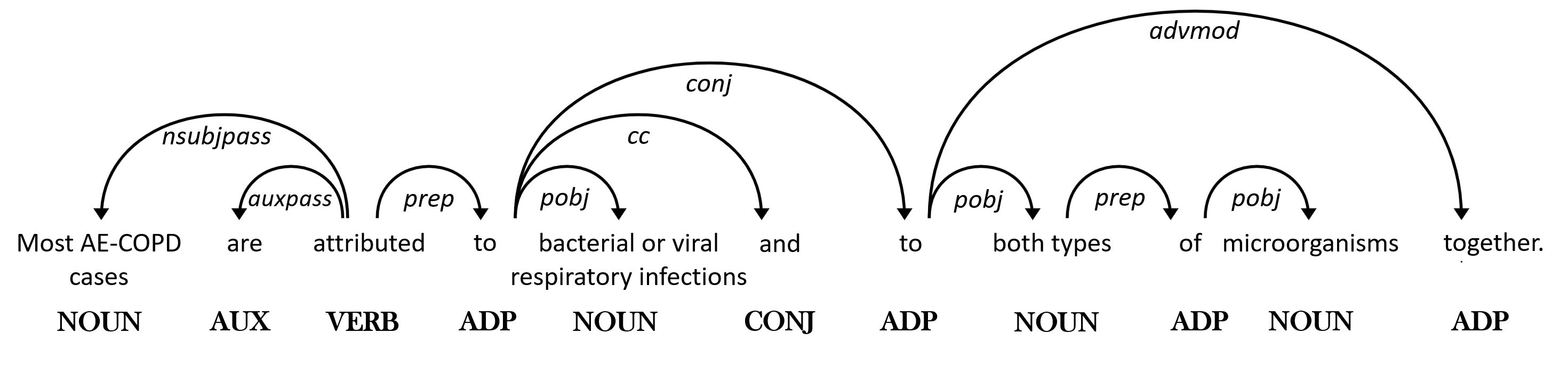}
    \caption{Dependency tree example for a sentence}
    \label{fig:displacy}
\end{figure*}

Given a sentence \textit {``Most AE-COPD cases are attributed to bacterial or viral respiratory infections and to both types of microorganisms together''}, Figure \ref{fig:displacy} shows the dependency tree of this sentence parsed with Spacy. As can be
seen, the vertices of the dependency tree are the words or phrases labeled by
parts-of-speech; each edge reflects a dependency relation between the two words 
associated to the end-points of that edge.

%All the patterns of $\mathcal{P}_C$ are identified by causal phrases, causal links, causative verbs and different cause-effect semantic  patterns.

\subsection{Dependency Pattern Extraction}
Dependency pattern is a linguistic structure which denotes a relationship between entities
in a sentence. For denoting cause-effect relationship, there exist several dependency 
patterns in English literature, and a comprehensive compilation of these patterns are 
needed for
extracting cause-effect entities with high recall. To obtain such a listing of cause-effect
dependency patterns, we use a supervised machine learning approach. Before discussing the
machine learning method, we provide a formal definition of dependency patterns as below.

\subsubsection{Definition of Dependency Pattern}
Generally, a pattern template of a semantic relation is associated with words or phrases which exhibit that relation in a sentence. For instance, phrases, such as, ``caused by'',
``attributed to'', etc.\ are phrases which are associated with cause-effect relationship
between a pair of entities in a sentence. Dependency patterns are formal representation
of such phrases through the use of dependency edges obtained from a dependency tree representation of a cause effect sentence.
For such a sentence, dependency edges are comprised of incoming and outgoing causal phrases, causative verbs, causal links, and their parts-of-speech (POS) tag. 

For instance, let us consider the sentence in Figure~\ref{fig:displacy}, the dependency edge ``attributed$~\rightarrow~$to'' is part of a dependency pattern because this edge is associated to ``attributed to'', which denotes a causal relation between the cause phrase ``bacterial or viral respiratory infection'' and the effect phrase ``Most AE-COPD cases''. Besides, the POS
tags of these words provide specific information about ``attributed'', and ``to'', for
them being ``verb'', and ``prep'' respectively; hence, the POS tags are also included in the 
dependency edge information. Note that, the word ``attribute'', in isolation, does not
exhibits any cause-effect pattern, whereas dependency edge ``attributed$~\rightarrow~$to''
is part of a dependency pattern. From this pattern, we can then  build a cause-effect pattern template, ``Y attributed to X'' where X and Y are cause and effect terms, respectively.

\subsubsection{Supervised Learning Approach for Dependency Pattern Extraction} \label{ASPER}
Cause-effect pattern extraction can be done manually by experts.
However, it is a labor intensive process; besides, the list of
patterns extracted by experts may not cover a large variety of
cause-effect sentences. In an earlier work,
we developed a supervised learning approach, called 
ASPER~\cite{kabir:2021:Asper}, for extracting syntactic patterns associated with any semantic relation. We utilize ASPER for collecting a larger number of cause-effect dependency patterns.
Note that, ASPER is used only to find the dependency patterns, but \name\
uses those patterns for extracting cause and effect phrases.
In the next paragraph we briefly illustrate the dependency pattern extraction task performed by ASPER.

\captionsetup[figure]{labelfont=blue}
\begin{figure*}[t]
    \centering
    \includegraphics[scale=0.55]{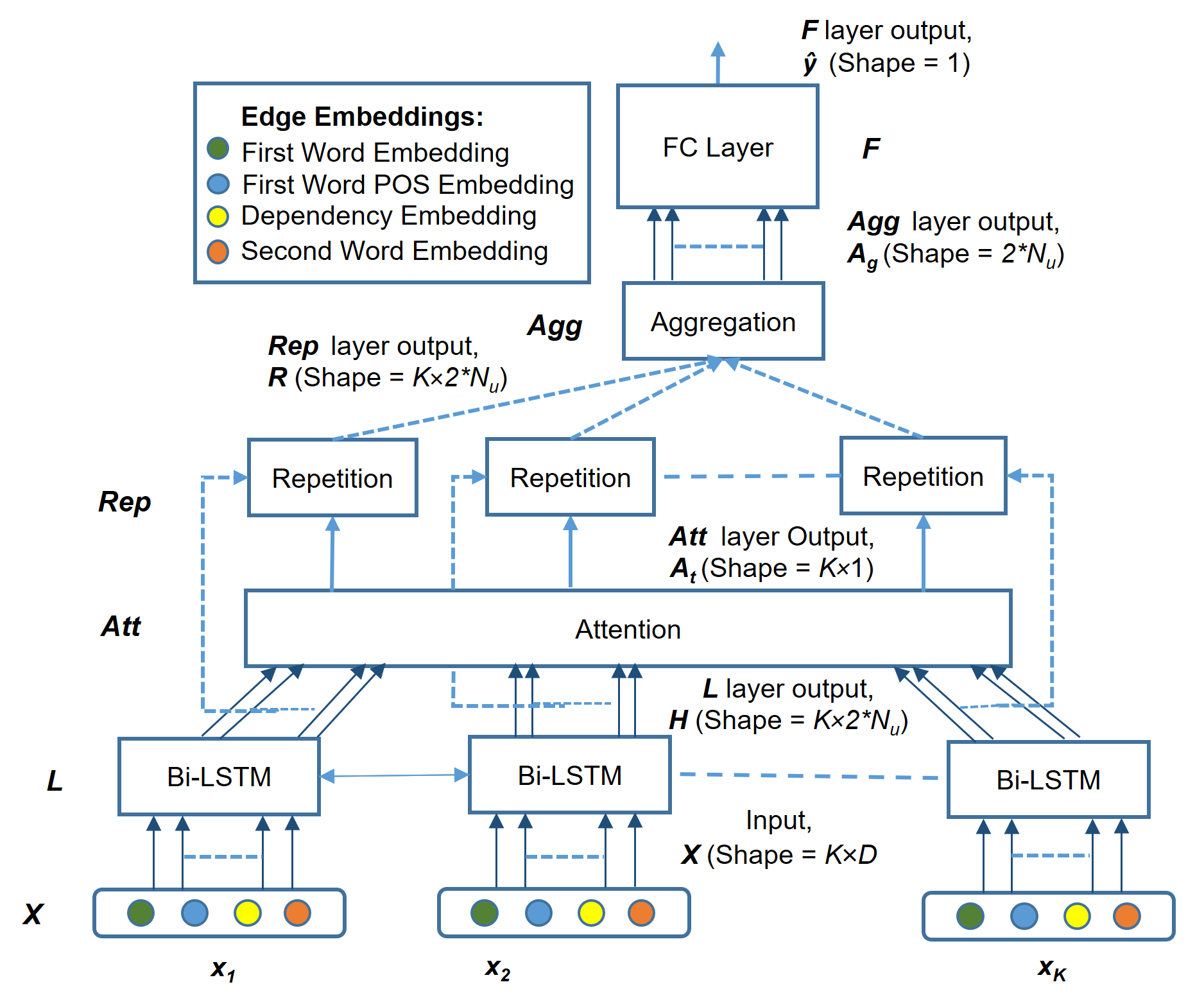}
    \caption{Description of LSTM with attention layer for binary semantic relationship classification with ASPER model.}
    \label{fig:classifier}
\end{figure*}

Dependency patterns are extracted from a collection of sentences exhibiting 
cause-effect relation between a cause-effect pair. Different labeled datasets are 
available in the literature which provide cause and effect terms in sentences, such as, ADE~\cite{Gurulingappa-2012-ADE}, Semeval-2007 \cite{girju-etal-2007-semeval}, and Semeval-2010~\cite{hendrickx-etal-2010-semeval}. Some term-pairs in these datasets do not contain cause-effect relation in their corresponding sentences. These pairs are referred as negative pairs. For our purpose, ASPER in Figure~\ref{fig:classifier} is trained to solve a binary classification
task to predict whether there is a cause-effect relationship between a given term-pair in a given sentence. The input to ASPER is a
collection of edges in the dependency tree representation of the given sentence. The edges are ordered as per shortest path from the cause node to effect node in the dependency tree.

After that, an attention based bi-directional LSTM model is trained for binary classification. The edges which are important with respect to attention values are then collected to perform frequent itemset mining~\cite{Zaki:2000:Eclat} to collect patterns. Some sentences can exist in a dataset which possess cause-effect relation but lacks a strong pattern or any pattern at all. In that case, the edges which contain the cause or effect terms get more attention and also frequent itemset mining does not extract any pattern because of infrequency. Below, we provide a formal discussion of the extraction of dependency patterns.

Given a sentence $\mathcal{S} = [w_1, w_2, w_3 ....  w_N]$ with $N$ words or phrases  $w_1, w_2 ....  w_N$, let  $(u,v)$, where $u, v \in \{w_i\}_{\{1 \le i \le N\}}$ 
is a pair of noun phrases exhibiting cause effect relation. If 
$\mathcal{S}$ is parsed with a dependency tree parser, $\mathcal{T} = (V, E)$, each vertex, $v_{i} \in V$ is associated with a word $w_i$; besides,
the vertices are labeled with the parts-of-speech (POS) tag 
(e.g. noun, verb, adverb, adjective, etc.) of $w_i$. 
The edges-set, $E$ is the set of all directed edges in the dependency tree. Each dependency edge $e_{mn}$ links a parent vertex $v_{m}$ to a child vertex $v_{n}$. We describe the edge $e_{mn}$ as
$[w_m, pos(w_m), dep_{mn}, w_n]$, where $w_m$ and $w_n$ are words or phrases
associated with $v_m$ and $v_n$, $pos(v_m)$ denotes the parts of speech tag of $w_m$ and $dep_{mn}$ symbolises the dependency relation between $v_m$ and $v_n$. For sentence embedding, among all the edges of $E$, only the shortest path edges between $u$, and $v$ are considered. To explain the embedding of any edge $e_{mn}$; $w_m$, and $w_n$ are embedded with semantic embedding method, whereas $pos(w_m)$, and $dep_{mn}$ are embedded with one hot embedding. All of these embeddings are then concatenated for edge embedding.

Let for $\mathcal{S}$, there contains $K$ edges, which are embedded as $x_1$, $x_2$, ... $x_K$. The Bi-LSTM layer of Figure~\ref{fig:classifier}, $\mathcal{L}$, takes $x_{i}$  as input and outputs two hidden state vectors. The first hidden state vector, $\overrightarrow{h_i}$, is the forward state output, and the second hidden state vector, $\overleftarrow{h_i}$, is the backward state output. Let ${h_i}$ be the concatenated output of  
$\overrightarrow{h_i}$ and $\overleftarrow{h_i}$. Also, we define $\mathbf{H}$, which is the concatenation of 
each $h_i$ output from $\mathcal{L}$ for $x_i$.

\[ \overrightarrow{h}_{i} = \mathcal{L}(\overrightarrow{h}_{i-1},x_{i}),  \overleftarrow{h}_{i} = \mathcal{L}(\overleftarrow{h}_{i+1},x_{i})\]
\[ {h}_{i} = [\overrightarrow{h}_{i}, \overleftarrow{h}_{i}], \mathbf{H} = [{h}_{1}, {h}_{2}....{h}_{K}]\]

Following the Bi-LSTM layer, $\mathcal{L}$, output $\mathbf{H}$ is used as input to the attention layer, $\bf{Att}$. The attention layer produces, $\mathbf{A}_{t}$, a vector of size $K \times 1$ where each $a_{i} \in \mathbf{A}_{t}$ is a value within a fixed range, $a_{i} \in [0, 1]$. Each such attention value, $a_{i}$, will encode the relative importance of edge embedding $x_{i}$ in making the binary classification decision. $\mathbf{A}_{t}$ is computed as below.

\[\it{Temp} = \bf{Tanh}( \mathbf{H} * \bf{W_{1}} ) * \bf{W_{2}} \]
\[\mathbf{A}_{t} = \bf{Softmax}(\it{Temp})\]

Here $\bf{W_{1}}$ is a trainable matrix of shape $2*N_{u} \times 2*N_{u}$, $\bf{W_{2}}$ is another trainable matrix of shape $2*N_{u} \times 1$. The shape of temporary variable $\it{Temp}$ is $K$, on which we 
apply $\bf{Softmax}$ activation to retrieve $\mathbf{A}_{t}$.

Next, the model uses both $\mathbf{A}_{t}$ and $\mathbf{H}$ as inputs for the repetition layer, $\bf{Rep}$. The repetition layer, $\bf{Rep}$, outputs $\mathbf{R}$ of shape $K \times 2*N_{u}$. $\mathbf{R}$ is simply the scalar multiplication of each hidden input $h_{i} \in \mathbf{H}$ with its corresponding scalar attention value, $a_{i} \in \mathbf{A}_{t}$.

Then, the model uses $\mathbf{R}$ as input for the aggregation layer, $\bf{Agg}$. The aggregation layer simply computes the column-wise sum of $\mathbf{R}$ in order to yield the $2*N_{u}$ shape output, $\mathbf{A}_{g}$. In short, $\mathbf{A}_{g}$ outputs the weighted sum of $\mathbf{H}$ where weights are the attention values. 

\[\mathbf{A}_{g} = \bf{Summation}(\mathbf{R})\]

$\mathbf{A}_{g}$ is then used as input to a fully-connected layer with sigmoid activation function, whose output is a scalar, $\hat{y}$, which denotes the prediction of a binary label, $y$.

\[\hat{y} = \bf{Sigmoid}(\bf\mathbf{A}_{g} * {W_{3}})\]

Here $\bf{W_{3}}$ is a randomly initialized weight matrix of shape $2*N_{u} \times 1$.

Using these constructs, we train the binary classifier using the edge embeddings to predict whether the sentence $S$ exhibits cause-effect relation for a given cause-effect pair. We train the model using standard binary cross-entropy loss:
$Loss = -\frac{1}{|\mathcal{T}|} \sum_{t \in \mathcal{T}} y_{t} * log(\hat{y}_{t}) + (1 - y_{t}) * log(1 - \hat{y}_{t})$.
Using Early Stopping~\cite{Caruana:2001:EarlyStopping}, we train the model until the validation loss does not decrease at the end of an epoch and then load the model parameters of the previous epoch in which validation loss has decreased.

 While the model learns to classify, it identifies important edges based on attention values which contribute in classification. To make ASPER corpus independent, we introduce frequent itemset mining over the collected edges and extract the complete frequent dependency patterns. The statistics of the patterns and performance of ASPER can be found in our previous work~\cite{kabir:2021:Asper}.

\subsubsection{Validation of Dependency Patterns and Template Creation}
ASPER extracts cause-effect dependency pattern with high precision although partial and noisy patterns are extracted at times. Moreover, to ensure better recall for cause-effect pair extraction we need to work with strong dependency patterns. Additionally, template patterns are easily convertible to dependency patterns and the scope of this paper is causality extraction from a sentence, which can be started from the pattern templates as the templates are human recognizable patterns. The templates we provide can be modified if necessary for future work. Finally, filtering and validating extracted patterns are way easier than finding patterns from scratch. Below, we want to illustrate how we create templates from dependency pattern using an example.

Pattern template creation task is performed once dependency patterns are collected from ASPER. For an instance of dependency pattern, consider the example sentence in Fig~\ref{fig:displacy}. In this sentence, 
the cause phrase $X$ is \textit{``bacterial or viral respiratory infections''}, and the effect phrase 
$Y$ is  \textit{``Most AE-COPD cases''}. The following dependency pattern, $\rho$
\begin{equation*}
    \{(\mathbf{attribute},verb,nsubjpass,Y), (\mathbf{attribute},verb,prep,\mathbf{to}), (\mathbf{to},adp,pobj,X)\}
\end{equation*}
is extracted by ASPER where $X$ and $Y$ can be any cause and effect term respectively.
From this dependency pattern, we have introduced four pattern templates, \textit{Y attributed to X}, \textit{Y is attributed to X}, \textit{Y can be attributed to X}, and \textit{Y, which is attributed to X}. While there can be a lot more templates other than these, most of these template patterns lead to the same dependency pattern they come from. For example, if \textit{Y is attributed to X}, and \textit{Y, which is attributed to X} are parsed with a dependency parser, and the dependency edges in the shortest path between $X$, and $Y$ are observed, the edges will be identical to the edges of $\rho$. 

\subsubsection{Regenerating Dependency Patterns from Pattern Templates}
While most of the templates for a dependency pattern lead to the same dependency pattern, occasionally there can be some marginal changes in the dependency edges. Sometimes, POS tags are changed. For instance, for the sentence \textit{Malaria is caused by the Plasmodium parasite}, \textit{Malaria} is a \textit{PROPN}. On contrary, consider \textit{Most fractures are caused by a bad fall or automobile accident}. Here the effect term is \textit{Most fractures} which is \textit{NOUN}. To overcome this, we generate a lot of dummy sentences using the pattern templates, replacing X and Y with some actual cause-effect terms, for example (\textit{fire}, \textit{damage}), (\textit{Malaria}, \textit{Plasmodium parasite}) etc. Once the sentences are constructed, we parse those sentences with dependency parser. Then the shortest path edges are calculated with identical approach as ASPER described in \ref{ASPER}.
Actual cause-effect terms of those edges are replaced by $X$ and $Y$ for general dependency patterns. These final dependency patterns are stored in $\mathcal{P}_C$ for causality extraction. Column 1 of Figure~\ref{tab:pattern-result} shows some pattern templates, whereas column 3 shows the corresponding dependency patterns which are stored in $\mathcal{P}_C$. Note that, with the described method we have 142 dependency patterns stored in $\mathcal{P}_C$, and all of the patterns of $\mathcal{P}_C$ are used by \name\ for causality extraction.

\subsection{Pattern Matching and Extraction of Cause Effect Candidates}\label{pattern-matching}

For extracting cause and effect phrases from a test sentence, $\mathcal{L}$, we simply 
need to search whether the dependency tree of $\mathcal{L}$ has two nodes $u$ and $v$ 
such that the shortest path between $u$ and $v$ matches with any pattern $\mathcal{P}$ in 
$\mathcal{P}_C$. Obviously, we do not know which pair of nodes may qualify the above test;
we also do not know which pattern may appear in $\mathcal{L}$; so, we search over  
all possible pairs of nodes of $\mathcal{L}$'s dependency tree and all possible 
patterns. On some occasions, for a valid cause-effect relationship in $\mathcal{L}$, 
the match can be partial, i.e., only a subset of edges in the pattern may appear in 
the shortest path between $u$ and $v$. To positively recall such cases, we consider
a match to be acceptable only if a fraction (between 0.5 and 1) of pattern edges appear 
in the shortest path. This fraction is called $minThreshold$ and it remains as a 
user-defined parameter. If the matching is successful, the words associated with $u$ node
are considered as cause phrase candidates and the words associated with $v$ node are 
considered as effect phrase candidates. The experimental results that we show in this
paper are generated using $minThreshold$ equal to 1.

\subsection{Complete Phrases Extraction from the Candidates} \label{dependency}
For many sentences in scientific literature, the candidate cause and effect phrases are
not complete. So, we need to extend both the phrases, if such situation arises. To do that, we again
take cues from the dependency tree. Assume that for a sentence $\mathcal{L}$, its 
dependency tree nodes $u$ and $v$ are found to be cause and effect candidates. Say,
$w_u$ and $w_v$ are phrases of $\mathcal{L}$, which are associated to those two nodes, 
respectively. To extend the cause phrase we collect the ancestor and descendent nodes of $u$ in the dependency tree. 
We then create a phrase $w_a$ by concatenating the words associated with the ancestors of $u$ by maintaining their
order in $\mathcal{S}$. Likewise, all the successors of $w_u$ are used to find another phrase $w_s$. Finally, 
between $w_a$ and $w_s$, the phrase (let $w_m$) with the maximum number of words is considered 
as the extension of $w_u$. The extension, $w_m$ may contain part of pattern words, stop words etc, so $w_m$ is cleaned by removing such words to produce $w_{cause}$, which is our final cause phrase for the sentence $\mathcal{S}$. An identical process is applied for the node $v$ to obtain the final effect phrase $w_{effect}$.
The pair $(w_{cause}, W_{effect})$ is the extracted cause-effect phrase from the sentence $\mathcal{S}$. Note
that a sentence $S$ may have multiple cause-effect phrases for different pair of dependency tree nodes, in that
case all such pairs are returned.\\

Please refer to Fig~\ref{fig:displacy} for a complete example of phrase extraction. The given sentence is:
``Most AE-COPD cases are attributed to bacterial or viral respiratory infections and to both types of microorganisms together”, which is represented with the dependency tree in Figure~\ref{fig:displacy}. consider the node, $u$ corresponding to \textit{both types}, and $v$ corresponding to \textit{Most AE-COPD cases}. The shortest path between them is the following:

\begin{equation}
\begin{split}
    \{(attribute,verb,nsubjpass,v),(attribute,verb,prep,to),\\(to,adp,conj,to),(to,adp,pobj,u)\}
\end{split}
\end{equation}

A pattern exists in $\mathcal{P}_C$ with 100\% match for the pair (\textit{both types}, \textit{Most AE-COPD cases}). Also, for the following pair $u =$\textit{both types} and $v=$\textit{bacterial or viral
respiratory infection}, the shortest path between them is the following:

\begin{equation}
\begin{split}
    \{(attribute,verb,nsubjpass,v),(attribute,verb,prep,to),(to,adp,pobj,u)\}
\end{split}
\end{equation}

Another pattern exists in $\mathcal{P}_C$ which shows 100\% match with this pair. So, both of these pairs
are considered as candidate cause-effect phrases. Now, the ancestors of \textit{both types} is 
the singleton set \{\textit{to}\}, as the other ancestors are not adjacent with \textit{to} in the sentence $\mathcal{S}$. 
Successors of \textit{both types} is the set \{\textit{of, microorganisms}\}. After concatenation of the
ancestors and successors with \textit{both types}, we obtain, 
\textit{to both types of microorganisms}; After cleaning up, the word ``to'' is omitted (because it is a stop word) and the final cause phrase becomes \textit{both types of microorganisms}. 
The node \textit{Most AE-COPD cases} and also the node \textit{bacterial of viral respiratory infection} remains 
unchanged after the extension. Final results from this sentence are two cause-effect pairs:
first is (\textit{bacterial or viral respiratory infection, Most AE-COPD cases}), and the second
is (\textit{both types of microorganisms, Most AE-COPD cases}).

\begin{algorithm}[t]
\caption{Cause-Effect Phrase Extraction from Sentence}
\label{algo:extract_pattern}
\small
\SetAlgoLined\DontPrintSemicolon
    \SetKwFunction{algo}{($\mathcal{S}, \mathcal{P}_C, minThreshold$)}
    \SetKwProg{myalg}{Extract-Causal-Phrase}{}{}
    \myalg{\algo}
    {
        $\mathcal{C} = []$\;
        $\mathcal{C_4} = []$\;
        $\mathcal{T} = DependencyParser(\mathcal{S})$\;
        
        \For{$(u,v) \in \mathcal{S}$}
        {
            $P = ShortestPath(\mathcal{T}, u, v)$\;
            \For {$\mathcal{P} \in \mathcal{P}_C$}
            {
                $mRatio = \frac{|\mathcal{P} \cap P|}{|P|}$\;
                \If{$mRatio >= minThreshold$}
                {
                    $\mathcal{C_4}.append((u,v))$\;
                }
                
            }
        }
        \For{$(m,n) \in \mathcal{C_4}$}
        {
            $C = Descendants(m)$\;
            $u = MergeIfSequential(C)$\;
            $u = Clean(u)$\;
            $C = Descendants(n)$\;
            $v = MergeIfSequential(C)$\;
            $v = Clean(v)$\;
            $\mathcal{C}.append((u,v))$\;
        }
        
        $\Return \ \mathcal{C}$\;
    }
\hspace{-0.2in}
\end{algorithm}

We summarize the whole process with a pseudo-code in Algorithm~\ref{algo:extract_pattern}.
The method \textbf{Extract-Causal-Phrase} takes the sentence $\mathcal{S}$, the pattern
collection $\mathcal{P}_C$, and the minimum threshold, $minThreshold$ as parameters.
Given the sentence, first we find the dependency tree of $\mathcal{S}$ (Line 4). Then 
in the nested for loop (Line 5-12), for all
node pairs $u, v$ of $\mathcal{T}$ and for all patterns in $\mathcal{P}_C$,
we obtain the shortest path between $u$ and $v$, and check whether a significant fraction
of the edges in the shortest path overlap with the pattern edges. If yes, the pair $u$ and $v$
are stored in $\mathcal{C_4}$ as candidate cause-effect phrase pair. Then for all
phrase pairs in $\mathcal{C_4}$, we extend and clean them as needed (Line 15-22).
The overall complexity of the above method is quadratic with the number
of words in a sentence and linear with the number of patterns.

\section{Experimental design}
In this section we show experimental results to validate the performance of our proposed
method. For this we use two datasets, SemEval and MedCause, which are discussed in 
details in Section~\ref{sec:dataset}. Both the datasets contain sentences along with labeled cause and effect phrases and our objective is to extract these
phrases from each of the sentences in unsupervised manner. The annotation is used only
for evaluation. We compare the performance of \name\ with 
three competing methods, which are discussed in Section~\ref{sec:baselines}.
We use precision, recall, and $F_1$ as evaluation metric. In Section
~\ref{sec:eval}, we discuss how these metrics are computed in our experiments.

\textbf{\subsection{Datasets}}
\label{sec:dataset}

\noindent
\textbf{SemEval}: 
This is a well-used dataset, built by combining the SemEval 2007 Task 4 dataset ~\cite{girju-etal-2007-semeval} and the SemEval 2010 Task 8 dataset~\cite{hendrickx-etal-2010-semeval}. The SemEval datasets provide predefined positive and negative sentences with corresponding entity pairs. The datasets also include predefined train and test partitions. We made a validation partition by borrowing 
from train and test datasets through uniform sampling.
In total, there are 7545 sentences in the train, out of those 922 are positive sentences. The validation dataset contains 166 positive sentences out of 1332 sentences in total. Lastly, the test dataset contains 3060 total number of sentences, out of which 339 are positive sentences.
Among the cause-effect pairs, 90\% are single words, 8\% are double words and the rest are multi word nouns or noun phrases.\\

\captionsetup[figure]{labelfont=black}
\begin{figure}
    \centering
     \includegraphics[scale=0.6]{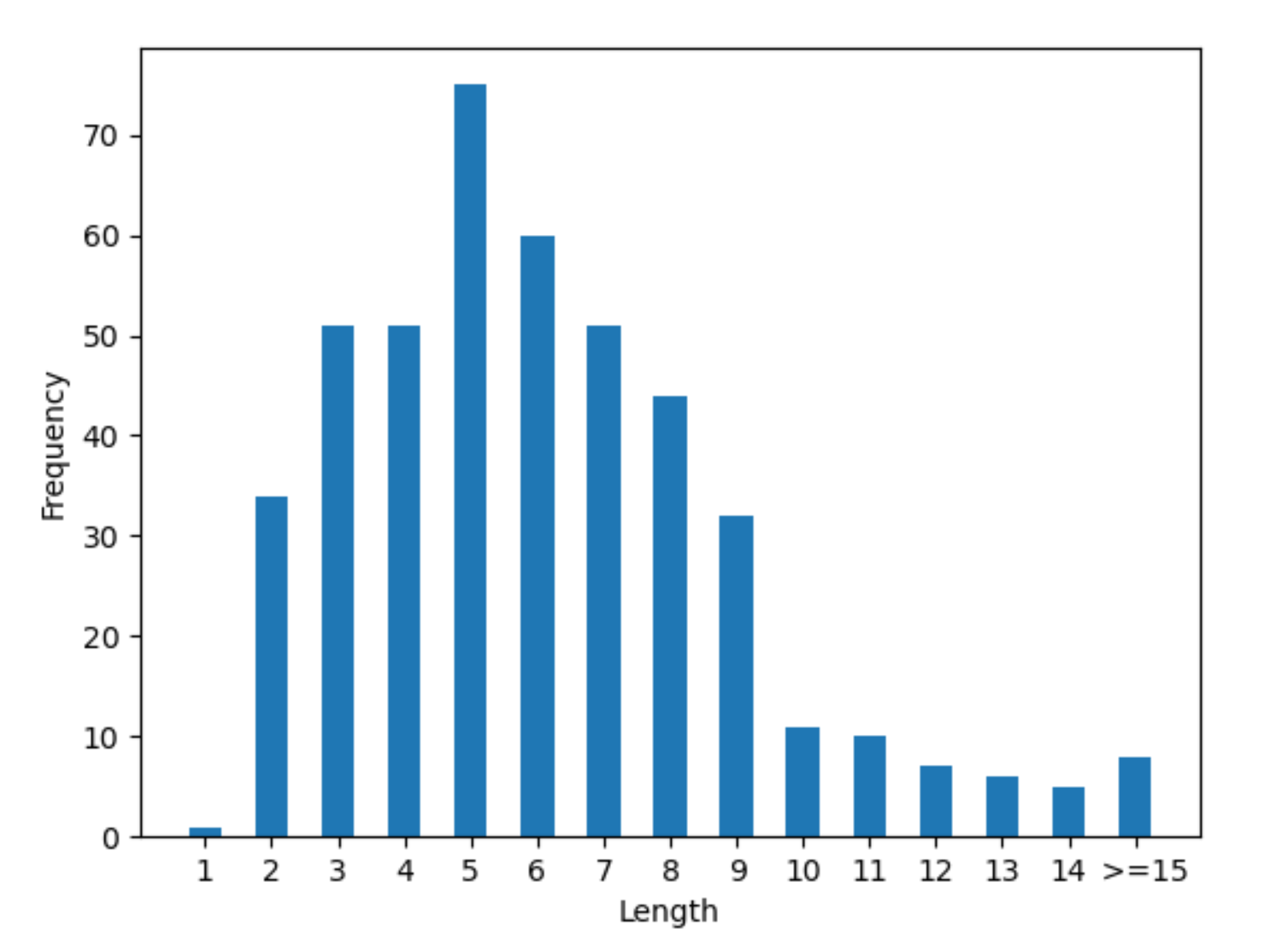}
    \caption{Histogram for \test\ dataset, where X-axis shows maximum number of words in a causal term, Y axis shows the frequency}
    \label{fig:histogram}
\end{figure}

\noindent
\textbf{MedCause}: We created this  dataset by  manual inspection of  a dump  of a large collection 
of PubMed articles. It contains 349 sentences, each of which are positive having confirmed 
cause-effect phrases. Three human experts labeled the cause and effect phrases from these sentences with
100\% inter-expert agreement. Note that, each sentence can contain multiple cause effect pairs, and
thus can contribute more than one cause-effect pairs for the dataset. After extracting cause-effect 
pair, we have 446 rows in the dataset. 

A key distinction of this dataset from SemEval is that in this dataset cause-effect phrases consist
of relatively larger number of words. To demonstrate this we draw a histogram showing the distribution
of word count in the cause or effect phrase (we took the largest of these two count) as shown in Figure~\ref{fig:histogram}. As can be seen the mode of the statistics is 5, and the majority of the
phrases have a count of higher than 5. On the other hand, more than 90\% of the cause or effect phrase 
in the SemEval dataset is single word, which makes the extraction of cause-effect phrases in the 
MedCause dataset much more difficult.\\

\textbf{\subsection{Competing Methods}} \label{sec:baselines} 
To show the comparative performance of \name\ we consider two competing methods. Out of these two
methods, Logical-Rule Based method is a hybrid method, having both unsupervised and supervised
components, and Word Vector Mapping Based method is purely unsupervised. More details of each of 
these methods are given below.\\

\noindent
\textbf{Logical-Rule Based}: This is a rule based hybrid method proposed by
Sorgente~\cite{Antonio:2013:baseline} where in the first step, a collection of cause-effect rules 
are used to extract cause effect candidates in an unsupervised manner. Unlike the dependency
patterns that we use in \name, these rules consist of different causative verbs in active or 
passive form, with or without preposition. These rules are matched in a given sentences to 
obtain cause and 
effect phrase candidates. However, not all the candidates they extract contain causal relationship. 
So, in a second step, they used a supervised binary classification to filter our false positive pairs.
To train the classification model, they use the train partition of SemEval dataset, and classify 
based on that trained model.

For our experiment, we also do the same; when reporting results on the SemEval dataset using 
Logical-Rule based method, we build the supervised part of the model by using the train and the 
validation partition of the SemEval dataset and then report results on the corresponding test dataset. 
For experiments on the MedCause dataset, the same model is used. We also build another version
of this classification model, which was trained with all SemEval train instances plus 20\% of the 
random data instances from the MedCause dataset. Intention of building this second model is to validate
whether retraining with sentences from the medical domain improves the performance of this method.

\noindent
\textbf{Word Vector Mapping Based:} This method is proposed for building a causal graph from a 
medical corpora, but this method can extract cause-effect terms as well~\cite{xiao:2019:graph}. 
It is an unsupervised method which uses regular expression based dependency parsing. Then pre-trained
Skip-Gram method of Word2Vec~\cite{word2vec-2013} is used to discover causative verbs with cosine 
similarity. From those causative verbs and regular expression based Parts of Speech parsing the
authors extract cause-effect terms from sentences. Extracted cause-effect terms are then used to 
form a causal graph. We use the causality extraction ability of this method and introduce it as 
one of the competing methods.\\

%The main innovation is 
%to discover causal words with Skip-Gram method. Those causal words or phrases are then used to 
%extract cause-effect term pairs. A dataset from the Asia Bayesian network and a dataset of the 
%risk factors of Alzheimer’s disease are constructed to validate this method in the original setup.

%\textbf{Dasgupta's method} We have introduced a bi-lstm based deep learning method to extract cause effect pairs as one of the  baselines.\cite{dasgupta-etal-2018-automatic} They annotate each word or phrase in a sentence as either a cause, effect, pattern term or none. The bi-lstm model predicts for each entity. Only the predicted cause and effect terms are stored. However, this method requires annotation. In our case, for train data, we use semeval which has already the cause and effect terms; the pattern tags are annotated from causative verbs. For labelling phrase, each word is labelled as same entity. This method is tested on the \test\ dataset.

\noindent
\textbf{Dependency Pattern Based variant of \name}: We create this baseline in which dependency patterns are used.
So, in terms of sentence representation this method is identical to \name. However, for extracting 
phrases, this method relies on the noun phrase 
extraction~\cite{noun-phrase-2016,noun-phrase-2010} methods whose implementation is 
available from Spacy~\cite{Spacy2}. A problem with Spacy's noun phrase extraction processes is 
that they often extract only a single word, instead of complete cause and effect phrases. In order 
to enhance the phrase extraction performance, we also use an advanced phrase extraction technique, 
namely PKE~\cite{boudin:2016:COLINGDEMO}. ~~~\S \\

Note that, PKE can be used with earlier baseline methods also to improve their phrase extraction capability, so we present results for a second variant of all the competing methods in which PKE 
is used for extracting phrases. So, in total we have 8 methods for which we show comparison results. 3 logical-rule based methods,
2 word vector mapping based method and finally, \name\ and two of its dependency tree based variants.

Besides the above methods, there are some supervised methods available in the literature which can 
extract causality\cite{dasgupta-etal-2018-automatic,bi-lstm-crf-2019}; however, we do not 
consider these methods for comparison. This is due to the fact that their performance is highly 
dependent on the datasets on which they are trained on. For our task of extracting cause-effect 
phrases from scientific literature or medical domain, there are no annotated corpus available 
for supervised training, a fact corroborated by other researchers\cite{xiao:2019:graph}.
we do have the options of training these models using SemEval datasets, which we tried; 
but such a trained model performs very poorly on MedCause sentences due to highly different distribution data between \test\ and SemEval corpus. Being the fact that our method is 
unsupervised, we limited our comparison to the above three methods, for which the phrase 
extraction part is unsupervised.\\

\begin{table}[!t]
\caption{Comparison with Baseline Methods on \test\ dataset when $minSim = 100\%$}\setlength{\tabcolsep}{1.5pt}
        \centering
        \renewcommand{\arraystretch}{1.4}
     
            \scalebox{1}{
                \begin{tabular}{l|  c|  c|  c}
                    %\multicolumn{4}{c}{\bf Hyponym-Hypernym Classification Comparison} \\
                    \hline
                    \bf Method & \bf Prec & \bf Rec & \bf F$_1$ \\\hline
                    Logical-Rule Based~\cite{sorgente2013automatic} & 0.020  & 0.038 & 0.026 \\
                    Logical-Rule Based + PKE & 0.021 & 0.038 & 0.026 \\
                    Enhanced Logical-Rule Based + PKE & 0.024 & 0.039 & 0.030 \\
                     \hline
                    Word Vector Mapping Based~\cite{xiao:2019:graph} & 0.028 & 0.041 & 0.033\\
                    Word Vector Mapping Based + PKE & 0.031 & 0.049 & 0.038\\
                    %Dasgupta's Method & 0.122  & 0.029 & 0.047 \\
                     \hline
                    Dependency Pattern Based  & 0.142  & 0.135 & 0.138 \\
                    Dependency Pattern Based  + PKE & 0.512  & 0.428 & 0.520 \\
                    \bf PatternCausality & \bf 0.556  & \bf 0.530 & \bf 0.543 \\\hline
                \end{tabular}
            }
            \label{table:classification-results-100}
\end{table}
\begin{table}
   \caption{Comparison with Baseline Methods on \test\ dataset when $minSim = 80\%$}  
   \setlength{\tabcolsep}{1.5pt}
        \centering
            \renewcommand{\arraystretch}{1.4}
            \scalebox{1}{
                \begin{tabular}{l | c | c|  c}
                    %\multicolumn{4}{c}{\bf Hyponym-Hypernym Classification Comparison} \\
                    \hline
                    \bf Method & \bf Prec & \bf Rec & \bf F$_1$ \\\hline
                     Logical-Rule Based~\cite{sorgente2013automatic} & 0.044  & 0.110 & 0.063 \\
                     Logical-Rule Based + PKE & 0.044  & 0.112 & 0.063 \\
                     Enhanced Logical-Rule Based + PKE & 0.048  & 0.118 & 0.068 \\
                      \hline
                     Word Vector Mapping Based& 0.055 & 0.148 & 0.080\\
                     Word Vector Mapping Based + PKE & 0.062 & 0.152 & 0.088\\[5pt]
                     %Dasgupta's Method & 0.226  & 0.067 & 0.104 \\
                      \hline
                     Dependency Pattern Based  & 0.215  & 0.211 & 0.212 \\
                     Dependency Pattern Based  + PKE & 0.592  & 0.513 & 0.560 \\
                    \bf PatternCausality & \bf 0.609  & \bf 0.589 & \bf 0.600 \\\hline
                \end{tabular}
            }

        \label{table:classification-results-80}
\end{table}

\textbf{\subsection{Evaluation Metrics}}\label{sec:eval} 

For evaluation we use traditional binary classification metrics, such as
precision, recall, and $F_1$-score over the sentences.
However, as complete phrase extraction is a difficult task, we define Levenshtein 
similarity based evaluation metrics which allow some margins of freedom to each 
method in terms of phrase extraction. In this subsection we want to define all 
the evaluation metrics, and performance of the methods will be discussed in 
the later subsection.

For a sentence $\mathcal{S}$, let ($x$, $y$) be a cause-effect term pair predicted by any of the described methods. As there can be multiple cause effect pairs per sentence, we define $\mathcal{C}_p$ as a set containing all cause-effect-sentence triplets, ($x$, $y$, $\mathcal{S}$).  Similarly, let $\mathcal{C}_t$ be a set for test triplets. Overall precision and recall are then defined by the following equations.

\begin{equation} \label{eqn-prec}
    Prec = \frac{|\mathcal{C}_p\cap\mathcal{C}_t|}{|\mathcal{C}_p|}
\end{equation}

\begin{equation} \label{eqn-recall}
    Rec = \frac{|\mathcal{C}_p\cap\mathcal{C}_t|}{|\mathcal{C}_t|}
\end{equation}

Extracting a phrase from a sentence is not an absolute task, it depends on the perspective of 
a viewer. So, we have introduced another measure of precision and recall with Levenshtein similarity ratio which we call edit similarity ratio. Let two phrases be $s_1$ and $s_2$.
If only insertion, deletion, or substitution of a character is allowed to convert one phrase to another;  let $\lambda$ be minimum number of such operations needed for this conversion. If |$s_1$|, and |$s_2$| are total number of characters in $s_1$, and $s_2$ respectively;  then Levenshtein similarity ratio, $ePer$ is defined by equation~\ref{eqn-eper}. $ePer$ similarity gives a value between 0 and 100, so we use a threshold parameter,
$minSim$, on the similarity value which denotes the  minimum similarity necessary for two phrases for being similar. Thus when comparing a predicted cause-effect phrase with a ground truth cause-effect phrase, if $ePer >= minSim$, we count the prediction as correct. 

\begin{equation} \label{eqn-eper}
    ePer = \frac{max(|s_1|, |s_2|) - \lambda}{max(|s_1|, |s_2|)} \times 100\%
\end{equation}

\section{Results}
In  this  section,  we  want to  provide  an  extensive  evaluation of  \name\ on  two  benchmark datasets described before. We report precision, recall, and $F_1$ scores for both evaluation types(with and without $minSim$; without $minSim$ means $minSim$ threshold is set to 100\%). We also present our results by grouping the phrases based on the number of words in the phrases
to compare each method's ability of extracting longer medical phrases. However, such results 
are provided only for \test\ dataset as ninety percents of cause or effect phrase of SemEval datasets are single word, as noted in Section~\ref{sec:dataset}.

\textbf{\subsection{Comparison with Competing Methods on \test\ dataset}}

First, we present comparison results of different cause-effect phrase extraction methods 
on the \test\ dataset using precision, recall, and $F_1$ metrics. 
Table~\ref{table:classification-results-100} shows the results
considering exact match $(minSim = 100\%)$, whereas Table~\ref{table:classification-results-80}
shows the results for 80\% or more edit similarity $(minSim = 80\%)$. As we can see
a total of eight methods are shows, which are grouped (groups are separated by horizontal 
line) based on their methodologies. Our proposed method and its variants are in the last group.

Among all the methods, \name\ and its variants, which use syntactic dependency patterns 
perform substantially better than both Logical-Rule Based and Word Vector Mapping Based methods.
As we can see in the tables, for exact match and partial match, \name's $F_1$ values
are respectively, 0.543 and 0.600, whereas the best value for the same among the competing 
groups are 0.038 and 0.088. Clearly \name's performance is at least one order of magnitude 
better than the best of the methods in the competing groups. In fact, the performance of 
other dependency pattern based methods that we have proposed as baseline, though worse
than \name, is better than Logical-rule based or Word vector mapping based methods by
nearly one order of magnitude. For example, Dependency Pattern based + PKE has $F_1$ values 
of 0.520 and 0.560, which is the second best result overall after \name. These results
illustrate that dependency pattern based approach which we propose is much superior than 
the existing approach for extracting cause-effect phrases.

All of the dependency pattern based approaches recognize the cause and effect nodes in 
the dependency tree using patterns. But, they differ in the way they extract the phrases.
One of the baseline, Dependency pattern + PKE method uses PKE (phrase keyword 
extraction)~\cite{boudin:2016:COLINGDEMO} for phrase extraction. On the other hand,
\name\ uses a custom phrase extraction process for capturing longer phrases, which makes it 
better than other dependency pattern based methods. In summary, two-fold contribution of 
\name, first using dependency patterns for identifying cause and phrase nodes, then 
innovative phrase extraction makes it the winner among all the methods
that we have shown in these tables.

\begin{table}
 \centering\caption{Performance of the methods in SemEval Dataset when minSim = 100\%}\setlength{\tabcolsep}{1.5pt}
 \renewcommand{\arraystretch}{1.4}
\begin{tabular}{l|  c|  c|  c}\hline
\bf Method & \bf Prec & \bf Rec & \bf F$_1$ \\\hline
Logical-Rule Based & 0.71  & 0.47 & 0.57 \\
Logical-Rule Based + PKE & 0.63  & 0.39 & 0.48 \\
 \hline
%Dasgupta's Method & 0.86  & 0.82 & 0.839 \\
Word Vector Mapping Based & 0.74  & 0.47 & 0.58 \\
Word Vector Mapping Based + PKE & 0.69  & 0.45 & 0.54 \\
 \hline
\bf{Dependency Pattern Based}  & \bf{0.75} & \bf{0.51} & \bf{0.61} \\
Dependency Pattern Based + PKE & 0.7 & 0.47 & 0.56 \\
\name\ & 0.71 & 0.48 & 0.57 \\\hline
\end{tabular}
\label{table:semeval-results}
\end{table}

The poorest performer among all the methods are Logical-Rule based methods. In fact, logical
rules cannot identify cause-effect terms well; this is because, the rules to extract causal 
terms are not adequate, and the rules are mainly logical rules which are not aware of syntactic structure of a sentence. Even if we apply PKE for phrase extraction, such a method still suffers. 
Note that, in the basic Logical-rule based method (Row 1), rules are obtained
from the SemEval dataset; so in the Enhanced Logical-rule based method + PKE (Row 3), 
we borrowed 20\% data from \test\ to SemEval in anticipation of getting better rules, yet 
the performance hardly improved.

Word vector mapping based methods performs better than Logical-Rule based methods.
Such methods, although find the causative verbs with cosine similarity, fails to
capture the syntactic structure of a sentence. Moreover, not all the
causative verbs are equally effective for exhibiting cause-effect relation. Another reason
for poor performance for this method is that it does not have any specific phrase extraction 
technique which is needed for \test\ dataset where cause-effect phrases are relatively longer.
We tried to enhance this method with PKE based phrase extraction, which improved its performance
noticeably, as can be seen from Table~\ref{table:classification-results-100} and
~\ref{table:classification-results-80}; yet, the improved performance is substantially
poorer than \name\ and its other variants.

\textbf{\subsection{Comparison with Competing Methods on SemEval dataset}}

In Table~\ref{table:semeval-results} we show results of out methods and other competing
methods on SemEval dataset for exact match scenario. As most of the cause-effect terms 
are single word nouns in this datasets, all methods performs much better on this dataset
and their performance are somewhat similar. The best performance is shown by 
Dependency pattern based method, one of \name\ variant. Its $F_1$-score is 0.61, whereas
best among Word vector based and Logical-rule based method is 0.58 and 0.57, respectively.
This validates that dependency pattern that we propose in \name\ is the best tool for 
extracting cause and effect phrase even for single word scenarios.
Interestingly, in all these methods, using dedicated phrase extraction tools, like PKE or 
the one that \name\ uses, makes the result worse. This is due to the fact that PKE or other
phrase extraction method tries to make the cause and effect phrase longer, but mostly all 
the cause-effect phrases in this dataset is of single-word length.

\begin{table}
\caption{Performance of all the methods in \test\ dataset for most densely populated lengths}
\setlength{\tabcolsep}{1.5pt}\footnotesize
\renewcommand{\arraystretch}{1.4}
\begin{tabular}{l|c|c|c|c|c|c|c|c}
%\multicolumn{9}{c}{\bf F1 Scores per Grams} \\
\hline
\multirow{2}{*}{\bf Method} & \multicolumn{8}{c}{\bf Length of Cause of Effect Phrase}\\
\cline{2-9}
 & 2 & 3 & 4 & 5 & 6 & 7 & 8 & 9 \\
\hline
Word Vector Mapping Based  &   0.317 & 0.161 & 0.001 & 0 & 0 & 0 & 0 & 0 \\
 %   &   &  &  &  &  &  &  & \\[6pt]

Word Vector Mapping + PKE &   0.330 & 0.202 & 0.021 & 0.013 & 0.016 & 0 & 0 & 0 \\
%   &   &  &  &  &  &  &  & \\[6pt]
 \hline
Logical-Rule Based &   0.281 & 0.125 & 0.001 & 0 & 0 & 0 & 0 & 0 \\
% &   &  &  &  &  &  &  & \\[6pt]

Logical-Rule Based + PKE  &   0.312 & 0.140 & 0.023 & 0.015 & 0.024 & 0 & 0 & 0 \\
% &   &  &  &  &  &  &  & \\[6pt]

Enhanced Logical-Rule Based + PKE &   0.310 & 0.160 & 0.031 & 0.025 & 0.031 & 0 & 0 & 0 \\
%  &   &  &  &  &  &  &  & \\[6pt]
\hline

Dependency Pattern Based  &   0.482 & 0.173 & 0.031 & 0 & 0 & 0 & 0 & 0 \\
% &   &  &  &  &  &  &  & \\[6pt]

Dependency Pattern + PKE &   0.481 & 0.193 & 0.062 & 0.030 & 0.022 & 0.010 & 0 & 0 \\
%  &   &  &  &  &  &  &  & \\[6pt]
\bf{PatternCausality}   &   \bf 0.562 & \bf 0.557 & \bf 0.543 & \bf 0.463 & \bf 0.435 & \bf 0.422 & \bf 0.417 & \bf 0.414 \\
%  &   &  &  &  &  &  &  & \\
\hline
\end{tabular}
\label{gram-result}
\end{table}
\normalsize

\textbf{\subsection{Performance of the competing methods grouped by phrase length}}

The key contribution of our method is that it can extract longer cause-effect phrases,
whereas existing methods fail to do so. In this experiment, we validate that the
performance of existing methods increasingly become worse as the length of the phrase
increases. In Figure~\ref{fig:histogram}, we have shown that the length of the majority of
the causal terms in \test\ dataset is between 2 and 9. So, we partition the \test\ test
dataset based on the phrase length and then show the performance of each method on 
each of those partitions in Table~\ref{gram-result}. We can see that over all different 
lengths, \name\ has good performance. But for the competing methods, their performance 
drops significantly as the length increases. For several of the competing methods the 
performance drops to zero when the length of the phrases reaches five or more.

\begin{figure*}[h!]
    \centering
    \includegraphics[width=1.1\linewidth]{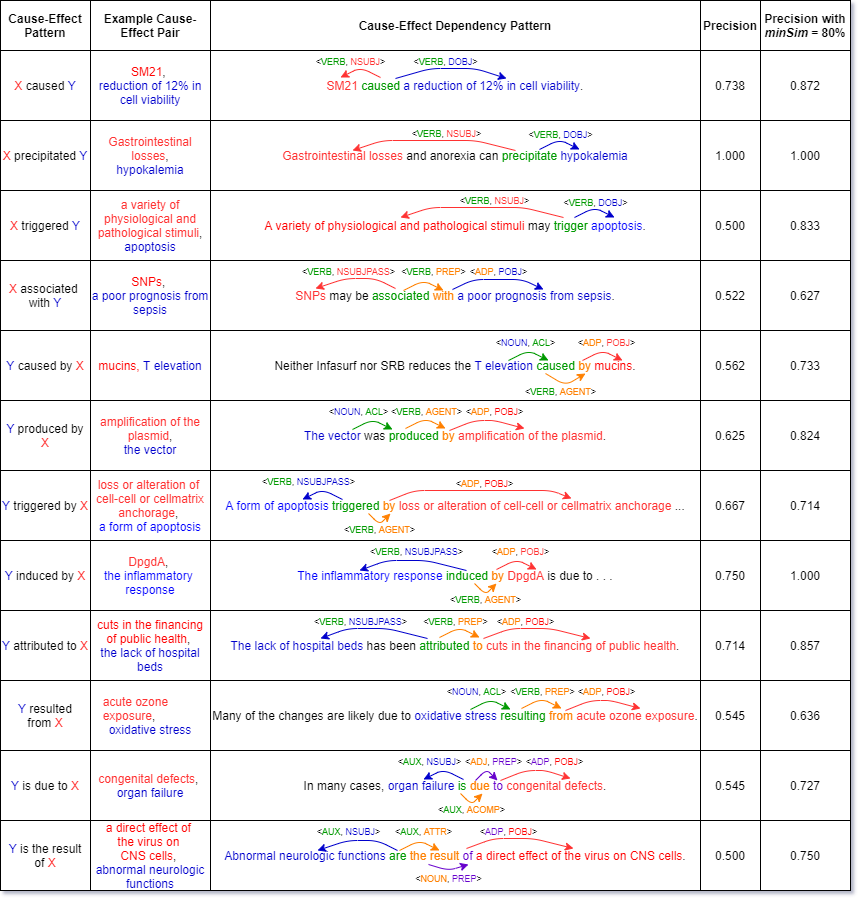}
    \caption{Cause effect pairs extraction result for some selected dependency patterns}
    \label{tab:pattern-result}
\end{figure*}

\textbf{\subsection{Performance of Selected Dependency Patterns}} \name's main contribution 
is to use dependency patterns to extract cause and effect phrases. To demonstrate the role of
dependency patterns, in Figure~\ref{tab:pattern-result}, we show a selected set of dependency
patterns along with an example sentence from the \test\ dataset, and its associated cause and 
effect phrases.
We also perform experiments to how precise a pattern is, i.e., how well a pattern can extract
the cause and effect phrases after it has been successfully used in a sentences. So, we
define a metric, named pattern precision, which defines the ratio of the number of correctly
predicted phrases over the total number of phrases predicted by a pattern. 

Precision of total 12 selected patterns with exact match and 80\% match are shown in 
Figure~\ref{tab:pattern-result}. Obviously the precision values are higher for the case of
partial (80\%) match than exact match scenario. The performance of rules are mixed; exact
match precision varying between .50 and 1.00. Note that, rules only identify the dependency
tree nodes associated to head words of cause and effect phrases, from where complete phrases
are extracted by using the phrase extraction method. So, precision of a rule is also affected
by the subsequent phrase extraction process. Most of the rules that we have used have precision 
better than 0.5.\\ 

%As no pattern can be absolute, there is no common ground for measuring recall of a pattern. But 
%precision of a pattern can be measured.

%80\% edit similarity although shows better result, it actually shows what best a pattern can perform. In the \test\ dataset there are total 28 patterns from $\mathcal{P}_C$, whereas SemEval dataset contains 18 of them.

\section{A Case Study}
Finally, we demonstrate the ability of \name\ to extract full cause and effect phrases by showing 
an example sentence from the \test\ dataset and analyzing how different methods perform on this 
sentence. We select the following sentence:
``Moreover,  amino acid sequence mutations in the new variant strains will cause  immunization failure 
of commercial vaccines.'' The causal phrase is ``amino acid sequence mutations in the new variant 
strains'', and the effect phrase is ``immunization failure of commercial vaccines''. \name\ extracts
both the cause and effect phrases exactly. Logical-Rule Based and other baseline approaches extract 
``the new variant strains'' as the cause phrase, and ``immunization failure'' as an effect 
phrase---incomplete phrases for both cause and effect. Dependency Pattern Based approach and Dependency Pattern + PKE extract ``amino acid sequence mutations'' and ``immunization failure'' as cause, and effect terms respectively, which are also incomplete.

\section{Conclusion and Future Work}

In medical domain,  causality extraction from literature is a very important task for knowledge
extraction, literature-based review, and hypotheses generation. But, existing cause-effect phrase
extraction methods are highly inadequate for solving this task with high accuracy. In most of the
cases, with existing methods, the extracted causality phrases are incomplete, which leads to knowledge
that to the best, is confusing, and to the worst, is inaccurate. Since, no existing methods pursue 
this task specifically for medical domain, we first created a manually annotated dataset, \test, which 
is the first  dataset of its kind. This is an important contribution towards the medical
information retrieval domain. Then we have contributed a novel method, 
\name, for causality extraction. Our proposed method is unsupervised, so it does not need large
annotated corpus for training, which makes it immediately usable. It is also extendable, as more
dependency patterns can be added to the pattern library to improve its performance.
Finally, we have demonstrate through detailed experiments that \name\ is highly
effective to extract long cause and effect phrases, whereas other competing methods fail to do so.
We also build other variants of \name, which uses only dependency patterns or uses a different
phrase extraction tool, namely PKE to demonstrate that dependency pattern
based cause-effect phrase extraction is an effective unsupervised approach.

In this work, we do not compare \name\ with any supervised approach, such as LSTM or other 
sequence-based model. The reason for that is lack of large datasets for the purpose of training.
So, one of the future goals is to first extract and annotate adequate sentences using \name\ 
and then train an effective supervised model for solving this task. Secondly, we have observed that the words considered for larger phrases are sequential in the Spacy dependency tree for majority cases. However, there are exceptions in this assumption. We want to extend \name\ in future to deal with those cases. Authors are committed to 
reproducible research and they will publish the \test\ dataset, code, and the dependency patterns,
once this paper is accepted.

\section{Conflict of Interest}
All authors declare that they have no conflicts of interest.

\bibliographystyle{spphys} 
\bibliography{sample}

\end{document}